\pdfoutput=1
\documentclass[letterpaper]{article}
\usepackage[margin=1.5in]{geometry}
\usepackage{setspace}
\usepackage{apacite}
\usepackage[T1]{fontenc}
\usepackage{enumitem}
\usepackage{amsmath}
\usepackage{tikz-dependency}
\usepackage{qtree}
\usepackage{gb4e}
\usepackage{amsthm}
\theoremstyle{definition}
\newtheorem{defn}{Definition}
\usepackage{bm}

\author{
  Eva Portelance (Stanford University)\thanks{Email: \texttt{portelan@stanford.edu}; Corresponding author}\\
  Amelia Bruno (McGill University)\\
    Daniel Harasim (\'Ecole Polytechnique F\'ed\'erale de Lausanne)\\
    Leon Bergen (University of California, San Diego)\\
    Timothy J. O'Donnell (McGill University)}
\title{Grammar Induction for Minimalist Grammars using Variational Bayesian Inference \\ \Large{A Technical Report}}
\date{\today}

\begin{document}

\maketitle

\begin{abstract} The following technical report presents a formal approach to probabilistic minimalist grammar parameter estimation. We describe a formalization of a minimalist grammar. We then present an algorithm for the application of variational Bayesian inference to this formalization.
  \\
  \textbf{Keywords:} grammar induction; probabilistic minimalist grammar; variational Bayesian inference; language modeling.  \end{abstract}

\section{Introduction}

In this paper we give an inference algorithm for minimalist grammars (MG) using variational Bayesian inference.  Variational Bayesian inference is an technique for inferring a distribution by formulating the inference problem as an optimization problem.  It can be contrasted with sampling-based approaches, and in many cases converges more quickly than those approaches.  Previous work that has applied the variational Bayesian approach to structured probabilistic models similar to our own include \citeA{kuriharasato04} for context-free grammars and \citeA{cohensmith10} for Adaptor grammars.\footnote{An even earlier application is \citeNP{mackay97}, which derives a variational Bayesian algorithm for Hidden Markov Models.}  There is work on other inference algorithms for grammars in the mildly context-sensitive space, such as combinatory categorial grammars \cite{biskhocken13, biskhocken12induct, biskhocken12simple, bisk15diss, wang16diss} and tree-adjoining grammars \cite{blunsocohn10}.  \citeA{hunterdyer13} formulate models and inference strategies for multiple context-free grammars, and show how they can be applied to MGs by exploiting the equivalence between the two formalisms.

\section{Minimalist Grammars} \label{MG}

Minimalist grammars are a formalization of \citeA{chomsk95}'s \textit{The Minimalist Program}, a theory of grammar which is the framework used for much of the current research in linguistic syntactic theory. The interest in studying these grammars is that they offer a direct line of comparison for resulting syntactic structure predictions within the linguistic minimalist literature.

The following formalization is based on \citeA{stable97, stable11}. A minimalist grammar is composed of a lexicon and structure building operations, typically classified as \textsc{merge} and \textsc{move}.
\paragraph{}

\begin{defn}
    A \emph{Minimalist Grammar} is a tuple $\langle \mathcal{L}, R\rangle$ where $\mathcal{L}$ is a finite set of lexical items and $R$ is a finite set of structure building operations. Each lexical item $l \in \mathcal{L}$ is a tuple $l = \langle \bm{\pi}, \bm{f} \rangle$ where $\bm{\pi}$ is a sequence of \emph{phonetic features} (in our case: a word or vocabulary item) and $\bm{f}$ is a sequence of \emph{syntactic features}. A tuple $\langle \bm{\pi}, \bm{f} \rangle$ is known as a \emph{chain} and lexical items are special cases of chains. Chains are denoted $\bm{\pi} : \bm{f}$. A sequence of chains is an \emph{expression}. Let $\mathcal{C}$ denote the set of chains and $\mathcal{C}^*$ the set of expressions. Every structure building operation $\rho \in R$ is a function from one or more expressions to an expression
    \[ \rho : (\mathcal{C}^*)^n  \to \mathcal{C}^*. \]
\end{defn}

\citeA{stable11} defines a variety of minimalist grammars which differ in (i) what the set of structure building operations $R$ contains, and (ii) what sorts of syntactic features are included in the lexicon. We will use a directional minimalist grammar as our reference MG.

\begin{defn}
    A \emph{Directional Minimalist Grammar} is a tuple $\langle \mathcal{L}, R\rangle$ where
    \begin{itemize}
        \item $\mathcal{L}$ is a finite set of lexical items whose syntactic features are of five types:
          \begin{enumerate}
            \item \emph{category} (e.g. \texttt{v, d, p}) - define the syntactic categories (verb, noun \ldots);
            \item right selector (e.g. \texttt{=d, =p}) - select argument constituent to the right;
            \item left selector (e.g. \texttt{d=, p=}) - selects argument constituent to the left;
            \item licensor (e.g. \texttt{+case, +wh}) - select moving constituent;
            \item licensee (e.g. \texttt{-case, -wh}) - selected moving constituent.
          \end{enumerate}
        \item $R = \{\text{merge}, \text{move}\}$ is the set of structure building operations. These are defined further below.
    \end{itemize}
\end{defn}

Each lexical item is categorized by one category feature, which  is to  say that there is exactly one category feature in the syntactic feature sequence of each lexical  item. If a lexical item has selector features, these are checked by merging with constituents of the corresponding category feature. A licensor feature is used to move a previously merged constituent with the corresponding licensee feature. We assume that all lexical item syntactic feature sequences appear in the following order:
\[ \text{(selector)^* (licensor)^* category (licensee)^*} \]

The two structure building operations are defined in \eqref{ex:ops} as deduction rules.
\begin{exe}
    \ex Let \texttt{x} be a category feature, $\gamma,\delta$ be non-empty sequences of syntactic features and $\lambda_1,\ldots,\lambda_m,\iota_1,\ldots,\iota_n$ be chains.

    \begin{xlist}
    \ex $\textsc{merge}: \mathcal{C}^* \times \mathcal{C}^* \to \mathcal{C}^*$ \label{ex:merge}
    \begin{xlist}
    \ex \textsc{merge}-L (merge a non-moving item to left)
    \[ \frac{[s:\texttt{x=}\gamma, \lambda_1,\ldots,\lambda_m] \quad [t:\texttt{x},\iota_1,\ldots,\iota_n]}{[ts:\gamma, \iota_1,\ldots,\iota_n,\lambda_1,\ldots,\lambda_m]} \]
    \ex  \textsc{merge}-R (merge non-moving item to right)
    \[ \frac{[s:\texttt{=x}\gamma, \lambda_1,\ldots,\lambda_m] \quad [t:\texttt{x},\iota_1,\ldots,\iota_n]}{[st:\gamma, \lambda_1,\ldots,\lambda_m,\iota_1,\ldots,\iota_n]} \]
    \ex  \textsc{merge}-m (merge an eventually moving item)
    \[ \frac{[s:\{\texttt{=x}\mid \texttt{x=}\}\gamma, \lambda_1,\ldots,\lambda_m] \quad [t:\texttt{x}\delta,\iota_1,\ldots,\iota_n]}{[s:\gamma, \lambda_1,\ldots,\lambda_m,t:\delta,\iota_1,\ldots,\iota_n]} \]
    \end{xlist}
    \ex $\textsc{move}: \mathcal{C}^* \to \mathcal{C}^*$  \label{ex:move}
    \begin{xlist}
    \ex \textsc{move}-1 (moving item to final landing position)
    \[\frac{[s:\texttt{+y}\gamma,t:\texttt{-y},\lambda_1,\ldots,\lambda_m]}{[ts:\gamma,\lambda_1,\ldots,\lambda_m]} \]
    \ex \textsc{move}-2 (moving item which will move again)
    \[\frac{[s:\texttt{+y}\gamma,t:\texttt{-y}\delta, \lambda_1,\ldots,\lambda_m]}{[s:\gamma,t:\delta,\lambda_1,\ldots,\lambda_m]} \]
    \end{xlist}
    \end{xlist}
    \label{ex:ops}
\end{exe}

In \eqref{ex:ops}, $\lambda_1,\ldots,\lambda_m,\iota_1,\ldots,\iota_n$ represent constituents which have previously merged using \textsc{merge}-m into the derivation, but still carry licensee features - i.e. items which have not reached their final landing position in the syntactic structure.

A \emph{derivation tree} represents a record of how a sentence is constructed. An example is given in figure 1, where we derive a sentence with a long distance dependency between a wh-word and the direct object of the verb.

\begin{figure} \label{fg:1}
\Tree [.{\textsc{move}\\\texttt{c}\\what did you see} [.{\textsc{merge}\\\texttt{+wh c, -wh}}
          [.{\texttt{=i +wh c}} ]
          [.{\textsc{merge}\\\texttt{i, -wh}\\(did you see, what)} [.{\texttt{=v i}\\did} ]
          [.{\textsc{merge}\\\texttt{v, -wh}\\(you see, what)}
            [.{\texttt{d -wh}\\what} ]
            [.{\textsc{merge}\\\texttt{=d v}\\(you see)}
              [.{\texttt{d= =d v}\\see} ]
              [.{\texttt{d}\\you} ] ] ] ] ] ]
\begin{enumerate}
 \setlength\itemsep{0em}
  \item \textsc{merge}-L: \texttt{v} selects \texttt{d} \textit{you};
  \item \textsc{merge}-m: \texttt{v} selects \texttt{d} \textit{what};
  \item \textsc{merge}-R: \texttt{i} selects \texttt{v} \textit{you see, what};
  \item \textsc{merge}-R: \texttt{c} selects \texttt{i} \textit{did you see, what};
  \item \textsc{move}: \texttt{-wh} moves to satisfy \texttt{+wh} \textit{what did you see};
  \item All features are satisfied.
    \caption{Derivation of 'what did you see?' in a MG $\langle \mathcal{L}, R\rangle$ where
  $\mathcal{L} $ = \{ what : \texttt{d -wh} , see : \texttt{d= =d v}, you : \texttt{d}, did : \texttt{=v i},  $\epsilon$ : \texttt{=i +wh c} \}
    }
\end{enumerate}

\end{figure}

Adding probabilities to a Minimalist Grammar is straightforward. Each lexical item is paired with a probability, which are normalized over lexical items with the same category feature. Given a MG $\langle \mathcal{L}, R\rangle$, let $C = \{c_1, \dots, c_K\}$ be the set of category features and for each $k$ where $1 \leq k \leq K$, let each $\bm{l}_k = \langle l_{k1}, \dots, l_{kM}\rangle$ contain $M$ lexical items having category feature $c_k$ in an arbitrary order, so that each $l_{km}$ is the $m$th lexical item with category feature $c_k$.

\begin{defn}
A \emph{Probabilistic Directional Minimalist Grammar} (PDMG) is a three tuple $\langle \mathcal{L}, R, \bm{\theta}\rangle$ where $\langle \mathcal{L}, R\rangle$ is an MG and $\bm{\theta} = \langle \bm{\theta}_1, \dots, \bm{\theta}_K \rangle$ are probability distributions over lexical items.

Recall that  $C = \{c_1, \dots, c_K\}$ is the set of category features found in $\mathcal{L}$. Let $\bm{l} = \langle \bm{l}_1, \dots, \bm{l}_K\rangle$ where each $\bm{l}_k$ is a sequence of the lexical items with category feature $c_k$, so that $\bm{l}_k = \langle l_{k1}, \dots, l_{kM}\rangle$ is the sequence of lexical items with the category feature $c_k$. The distributions over lexical items follow the same notational scheme: $\bm{\theta} = \langle \bm{\theta}_1, \dots, \bm{\theta}_K\rangle$ where each $\bm{\theta}_k = \langle \theta_{k1}, \dots, \theta_{kM}\rangle$ is a probability distribution over $\bm{l}_k$, such that
\begin{equation}
      P(l_{km}) = \theta_{km}
  \end{equation}
    Since each $\bm{\theta}_k$ is a probability distribution, they satisfy, for each $k$:
    \begin{equation}
        \sum_{m=1}^M \theta_{km} = 1
    \end{equation}
\end{defn}

\section{The generative model}
We will now define the formal representation we are interested in performing inference over. There exists a partial function from sequences of lexical items to well-formed derivation trees. A well-formed sequence corresponds to a depth first traversal of the order in which terminal nodes (lexical items) are merged into the derivation tree. In other words, first the root is listed, then its first child based on the selector features, then the children of that child and back up to the root's second child and so on, in polish notation such that the direction of child selectors does not matter. To guarantee that a sequence corresponds to a well-formed derivation tree, we use the following procedure, \textsc{is\_wellformed}, which takes a sequence of lexical items and return a boolean value. We assume the first lexical item in the sequence is the root of a tree (well-formed or not).
\begin{exe}
\ex \textsc{is\_wellformed} \\
Starting with the left most item in the sequence, move to the next item until you reach an item who first feature is a category feature:
\begin{enumerate}
\item If the first feature on the item is a selector feature, then move to the next item to the right and continue.
\item Else if the first feature on the current item is a category feature, then
\begin{enumerate}
    \item If this is the root item (left most at start) delete the feature and continue.
    \item Else if the first item to the left with a category feature (skipping items with only licensees) has a corresponding selector feature as its first feature (i.e. \texttt{=A} corresponds with \texttt{A}) delete both the category feature on the current item and the matching selector feature on the other item and continue from the current item.
    \item Else the tree is ill-formed.
\end{enumerate}
\item Else if the first feature on the current item is a licensee feature, then move back to the first item to its left.
\item Else if the first feature on the current item is a licensor feature, then find the first item to the right with a corresponding licensee feature (eg. licensee \texttt{-p} corresponds to licensor \texttt{+p}).
\begin{enumerate}
    \item If no such item is found then the tree is ill-formed.
    \item Else if any items with a category feature are in between the two items with the corresponding licensor/licensee features then the tree is ill-formed.
    \item Else delete the licensor feature on the current item and the corresponding licensee feature on the other item to its right and continue from the current item.
\end{enumerate}
\item Else if the current item has no more features, delete the item and move back to the item to its left if there is one, else to its right and continue.
\end{enumerate}
Continue this process until either the tree is found to be ill-formed or until all the items and their syntactic features have been deleted. If the sequence is empty at the end of this procedure, then the tree is well-formed and the procedure returns \textsc{true}, otherwise it returns \textsc{false}.
\end{exe}

For example, the sequence $\langle \epsilon$: \texttt{=i +wh c}, did:\texttt{=v i}, see:\texttt{d= =d v}, you:\texttt{d}, what:\texttt{d -wh}$\rangle$ corresponds to the well-formed derivation tree in figure 1. We can use the procedure described above to prove this.

\begin{exe}
    \ex \textsc{is\_wellformed}($\langle \epsilon$: \texttt{=i +wh c}, did:\texttt{=v i}, see:\texttt{d= =d v}, you:\texttt{d}, what:\texttt{d -wh}$\rangle$)\\
    Starting from the first item:
    \small{
\begin{enumerate}
    \item The first feature on the current item is the selector \texttt{=i}, so we move to the next item to its right, did:\texttt{=v i}.
    \item The first feature on the current item is the selector \texttt{=v}, so we move to the next item to its right, see:\texttt{d= =d v}.
    \item The first feature on the current item is the selector \texttt{d=}, so we move to the next item to its right, you:\texttt{d}.
    \item The first feature on the current item is the category feature \texttt{d}, so we find the first item to the left with a category feature, see:\texttt{d= =d v}, and confirm that the first feature on this item is the corresponding selector feature \texttt{d=} so we can delete both features and we are now left with the following sequence: $\langle \epsilon$: \texttt{=i +wh c}, did:\texttt{=v i}, see:\texttt{=d v}, you:\texttt{}, what:\texttt{d -wh}$\rangle$
    \item The current item, you: , has no more features so we can delete it and move back to the previous item to the left, see:\texttt{=d v}, and we are now left with the following sequence: $\langle \epsilon$: \texttt{=i +wh c}, did:\texttt{=v i}, see:\texttt{=d v}, what:\texttt{d -wh}$\rangle$
    \item The first feature on the current item is the selector \texttt{=d}, so we move to the next item to its right, what:\texttt{d -wh}.
    \item The first feature on the current item is the category feature \texttt{d}, so we find the first item to the left with a category feature, see:\texttt{=d v}, and confirm that the first feature on this item is the corresponding selector feature \texttt{=d} so we can delete both features and we are now left with the following sequence: $\langle \epsilon$: \texttt{=i +wh c}, did:\texttt{=v i}, see:\texttt{v}, what:\texttt{-wh}$\rangle$
    \item The first feature on the current item is the licensee \texttt{-wh}, so we move back to the previous item to the left the next item to its right, see:\texttt{v}.
    \item The first feature on the current item is the category feature \texttt{v}, so we find the first item to the left with a category feature, did:\texttt{=v i}, and confirm that the first feature on this item is the corresponding selector feature \texttt{=v} so we can delete both features and we are now left with the following sequence: $\langle \epsilon$: \texttt{=i +wh c}, did:\texttt{i}, see:\texttt{}, what:\texttt{-wh}$\rangle$
    The current item, see: , has no more features so we can delete it and move back to the previous item to the left, did:\texttt{i}, and we are now left with the following sequence: $\langle \epsilon$: \texttt{=i +wh c}, did:\texttt{i}, what:\texttt{-wh}$\rangle$
    \item The first feature on the current item is the category feature \texttt{i}, so we find the first item to the left with a category feature, $\epsilon$: \texttt{=i +wh c}, and confirm that the first feature on this item is the corresponding selector feature \texttt{=i} so we can delete both features and we are now left with the following sequence: $\langle \epsilon$: \texttt{+wh c}, did:\texttt{}, what:\texttt{-wh}$\rangle$
    \item The current item, did: , has no more features so we can delete it and move back to the previous item to the left, $\epsilon$: \texttt{+wh c}, and we are now left with the following sequence: $\langle \epsilon$: \texttt{+wh c}, what:\texttt{-wh}$\rangle$
    \item The first feature on the current item is the licensor feature \texttt{+wh}, so we find the first item to the left with a corresponding licensee feature, what:\texttt{d -wh}, and confirm that there are no items with a category feature intervening so we can delete both features and are left with the sequence: $\langle \epsilon$: \texttt{c}, what:\texttt{}$\rangle$
    \item The first feature on the current item is the category feature \texttt{c} and since this is the root item, we can delete it.
    \item The current item, $\epsilon$:,  has no more features so we can delete it and move to the next item, what: .
    \item This item, what: , is also out of features so we can delete it and we are left with an empty sequence -- confirming that the initial sequence is well-formed.
\end{enumerate}
}
\end{exe}

Given a sequence of lexical items $d = \langle l_1, \dots, l_n \rangle$,
we will say that the probability of and ill-formed sequence is 0, however, the probability of a well-formed derivation $d$ is simply be the product of the probabilities of the lexical items that generate it: \begin{equation}
   P(d) =
   \begin{cases}
   \prod_{i=1}^n P(l_i) &\text{\textsc{is\_wellformed}($d$)= \textsc{true}}\\
   0 &\text{\textsc{is\_wellformed}($d$)= \textsc{false}}
   \end{cases}
\end{equation}

For the rest of this technical report, we will assume that we are conditioning on all sequences $d$ being well-formed. For our generative model, we define Dirichlet priors over each $\bm{\theta}_k$, previously introduced in our PDMG, parameterized by
a sequence of prior pseudo-counts
$\bm{\alpha}_k = \langle \alpha_{k1}, \dots, \alpha_{kM}\rangle$.
\begin{equation}
  P(\pmb{\theta})=\prod_{k=1}^K \textsc{Dirichlet}(\pmb{\theta}_{k},\bm{\alpha}_k).
\end{equation}

Then our model is the following Dirichlet Categorical:
\begin{align}
    \bm{\theta}_k &\sim \textsc{Dirichlet}(\bm{\alpha}_k) \quad 1 \leq k \leq K
    \label{dirprior} \\
    l_{km} &\sim \textsc{Categorical}(\bm{\theta}_k) \quad 1 \leq m \leq M
    \label{catprior}
\end{align}
where $K$ is the number of categories in $C$ and $M$ is the respective number of lexical items in each $\bm{l}_k$.

So the joint probability of a well-formed derivation $d$ and parameters $\bm{\theta}$ is
\begin{align}
    P(d, \bm{\theta}|\bm{\alpha})
        &= P(d|\bm{\theta}, \bm{\alpha})P(\bm{\theta}|\bm{\alpha}) \\
        &= \prod_{i=1}^n P(l_{k_im_i} | \bm{\theta}_{k_i})P(\bm{\theta}|\bm{\alpha})
\end{align}

\section{Variational Bayesian Inference for Minimalist Grammars}
\paragraph{}

In Bayesian inference, we are interested in calculating the posterior probability of a latent variables given observed variables. Here, the observed data is a corpus of sentences and the latent variables are the best derivation trees over those sentences along with the probability vectors over the grammar. Let $S$ be our corpus, where $S = \langle s_{1}, \dots, s_{N} \rangle$, and let $D$ be the sequence of derivation sets for $S$, where $ D = \langle \bm{d}_{1}, \dots, \bm{d}_{N} \rangle $, such that $ \bm{d}_{n} = \{ d_{n1}, \dots, d_{nJ} \} $ is the set of $J$ possible derivations for $s_n$. Recall that $\bm{\theta} = \langle \bm{\theta}_k\rangle_{k=1}^K$ represents probability distributions over $\bm{l} =  \langle \bm{l}_k\rangle_{k=1}^K$, where each $\bm{l}_k$ is a sequence of lexical items of category $c_k \in C$, and that $P(l_{km}) = \theta_{km}$. Recall also that there is a Dirichlet prior over each $\bm{\theta}_k$, defined by the parameters $\bm{\alpha}_k = \langle \alpha_{km}\rangle_{m=1}^{M}$. We can now define the posterior we are interested in as follows:
\setcounter{equation}{\value{exx}}
\begin{equation}
  P(D, \pmb{\theta} | S, \bm{\alpha})=
  \frac{P(S | D, \pmb{\theta}, \bm{\alpha})P(D|\pmb{\theta},\bm{\alpha})P(\pmb{\theta}|\bm{\alpha})}{ \int_{\bm{\theta}}\sum_{D} P(S | D, \pmb{\theta}, \bm{\alpha})P(D|\pmb{\theta},\bm{\alpha})P(\pmb{\theta}|\bm{\alpha}) }
  \label{posterior}
\end{equation}

It should be noted that the prior is unnormalized because it is possible to randomly sample sequences of lexical items which do not form proper derivation tree as previously noted. That said, given that we condition only on well formed sequences of lexical items using the procedure described in the previous section the following updates are meaningful. Thus, We wish to find the assignments of probabilities to lexical items which maximize \eqref{posterior}.
The denominator of \eqref{posterior} is intractable to calculate exactly, so we must approximate the posterior. Here we show how to use variational Bayesian inference to do so. We give a very brief introduction to the technique, and then derive the variational distributions and variational updates needed to implement the technique for MGs.

Variational Inference (VI) was introduced by \citeA{attias99}, though the roots of the technique go back further to the technique called \emph{ensemble learning} in \citeA{hintonvancam93}. The derivation given in this section was to a large part based on the application of ensemble learning to HMMs in \citeA{mackay97} and the application of VI to CFGs in \citeA{kuriharasato04}.
VI formulates an inference problem as an optimization problem. If $P(D, \pmb{\theta} | S, \bm{\alpha})$ represents the true posterior, let $Q(D, \bm{\theta})$ represent our approximation. The Kullback-Leibler distance (KL) is a distance metric on probability distributions, so that $KL(P || Q) = 0$ iff $P = Q$. The KL distance is defined in  \eqref{kl}.
\begin{equation}
    KL(Q(\bm{z}) || P(\bm{z} | \bm{x})) =
    \mathbf{E}_{\bm{z} \sim Q}[\log Q(\bm{z})] -
    \mathbf{E}_{\bm{z} \sim Q}[\log P(\bm{z} | \bm{x})]
    \label{kl}
\end{equation}
where $\bm{z}$ are the latent variables and $\bm{x}$ are the observed variables.

In VI, the objective function we minimize is the KL-distance with respect to $Q$.
It can be shown that minimizing the KL-distance is equivalent to maximizing the Evidence Lower Bound (ELBO)\footnote{The ELBO is also known as the negative free energy, such as in \citeA{mackay97}. We do not give the proof of the equivalence of optimizing the KL-distance and the ELBO. For a good overview, see \citeA{bleietal17}}, defined in \eqref{elbo}.
\begin{equation}
    ELBO(Q(\bm{z})) =
    \mathbf{E}_{\bm{z} \sim Q}[\log P(\bm{z}, \bm{x})] -
    \mathbf{E}_{\bm{z} \sim Q}[\log Q(\bm{z})]
    \label{elbo}
\end{equation}
So the optimization we wish to approximate is
\begin{equation}
  \arg\min_{Q(D, \bm{\theta})} ELBO(Q(D, \bm{\theta}))
\end{equation}

In VI, we make independence assumptions for our variational approximations to simplify the optimization process. In particular, we assume that the variational distribution $Q$ can be broken up into independent distributions over the components of $D$ and $\bm{\theta}$ as such:
\begin{equation}
    Q(D, \bm{\theta}) = Q(D)Q(\bm{\theta}) =
    \prod_{n=1}^N Q(\bm{d}_n)
    \prod_{k=1}^K Q(\bm{\theta}_k)
\end{equation}

By making this simplifying assumption, we can optimize each component of $Q(D, \bm{\theta})$ separately.

In \S\S\ref{sec:qtheta}-\ref{sec:qd}, we show how to derive the optimal distributions $Q^*(D)$ and $Q^*(\bm{\theta})$. Then, in \S\ref{sec:updates}, we show how to use the $Q^*(D)$ and $Q^*(\bm{\theta})$ that we derived to formulate an iterative update algorithm for finding optimal parameters for the model.

\subsection{Optimization of $Q(\bm{\theta})$}
\label{sec:qtheta}

Expanding out the expectations, the ELBO of our distribution $Q(D, \bm{\theta})$ is
\begin{equation}
ELBO(Q(D, \bm{\theta})) = \int_{\bm{\theta}} \sum_{n \in N} Q(\bm{\theta},\bm{d}_n)
[\log P(s_n,\bm{d}_n,\bm{\theta}|\bm{\alpha}) - \log Q(\bm{d}_n,\bm{\theta}) ]
\label{ourelbo}
\end{equation}
where
\begin{equation}
\log P(s_n,\bm{d}_n,\bm{\theta}|\bm{\alpha}) = \sum_{k=1}^K \sum_{m=1}^M (\alpha_{km} - 1) \log \theta_{km}+
\sum_{j=1}^J \sum_{l_i \in d_{nj}} \log \theta_{k_{i}m_{i}} + c.
\label{fqnum}
\end{equation}
Here, we derive the optimal $Q(\bm{\theta})$. To do so, we wish to express \eqref{ourelbo}  as a function of $Q(\bm{\theta})$, keeping $Q(D)$ fixed. We restate \eqref{ourelbo} as \eqref{elboqtheta}, where the constant $c$ contains terms that depend only on $Q(D)$.

\begin{equation}
\int_{\bm{\theta}} Q(\bm{\theta}) \left[
\sum_{k=1}^K \sum_{m=1}^M (\alpha_{km} - 1) \log \theta_{km} +
\sum_{n \in N} Q(\bm{d}_n) \sum_{j=1}^J \sum_{l_i \in d_{nj}} \log \theta_{k_{i}m_{i}} -
\log Q(\bm{\theta})
\right] + c
\label{elboqtheta}
\end{equation}

If we represent \eqref{elboqtheta} in the form $\int_x Q(x) \log \frac{Q(x)}{P^*(x)}$, then Gibbs inequality implies that it can be minimized by setting $Q(x) = P^*(x)$. To this end, \eqref{elboqtheta} is equal to \eqref{elboqtheta2}.

\begin{equation}
 -\int_{\bm{\theta}} Q(\bm{\theta}) \log\left(\frac{Q(\bm{\theta})}{\prod_{k=1}^{K} \prod_{m=1}^{M} \theta_{km}^{\omega_{km}-1}}\right) + c
\label{elboqtheta2}
\end{equation}
where
\begin{equation}
\omega_{km} = \alpha_{km} + \sum_{n \in N} Q(\bm{d}_n) \sum_{j=1}^J \sum_{l_i \in d_{nj}} \delta(k_{i} = k, m_{i} = m)
\end{equation}
and where $\delta$ is the delta function
\[ \delta(\phi) = \begin{cases} 1, \phi \text{ is true} \\ 0, \text{otherwise}\end{cases}. \]
Then the optimized $Q*(\bm{\theta})$ is
\begin{equation}
    Q^*(\bm{\theta}) = \prod_{k=1}^{K} \prod_{m=1}^{M} \theta_{km}^{\omega_{km}-1}
    \label{qstar}
\end{equation}

Notice that this is a product over the formula for the probability density function of a Dirichlet. Accordingly, we can minimize \eqref{qstar} by choosing $Q^*(\bm{\theta})$ to be a product of Dirichlets:

\begin{equation}
Q^*(\bm{\theta}) = \prod_{k=1}^K \textsc{Dirichlet}(\bm{\theta}_k; \bm{\omega}_k)
\end{equation}
where $\bm{\omega}_k = \langle \omega_{k1}, \dots, \omega_{kM}\rangle$.

\subsection{Optimization of $Q(D)$}
\label{sec:qd}

We follow the same procedure as in \S\ref{sec:qtheta}. Express \eqref{ourelbo} as a function of $Q(D)$ with $Q(\bm{\theta})$ fixed.

\begin{equation}
    \sum_{n=1}^N Q(\bm{d}_n)
    \int_{\bm{\theta}} Q(\bm{\theta}) \left[
        \sum_{j=1}^J \sum_{l_i \in d_{nj}} \log \theta_{k_{i}m_{i}}
        - \log Q(\bm{d}_n)
    \right]
    + c
    \label{fqd}
\end{equation}
To represent \eqref{fqd} in the form $\int_x Q(x) \log\frac{Q(x)}{P^*(x)}$, we rewrite it as:
\begin{equation}
    -\sum_{n=1}^N Q(\bm{d}_n)
    \log \left[
        \frac{Q(\bm{d}_n)}{\prod_{j=1}^J \prod_{l_i \in d_{nj}} \theta^*_{k_im_i}}
    \right]
    + c
    \label{fdtheta}
\end{equation}
where
\begin{equation}
    \theta^*_{k_{i}m_i} =
    \text{exp}(\int_{\bm{\theta}} Q(\bm{\theta})\log \theta_{k_im_i})
\label{thetastar}
\end{equation}
Then the optimized distribution $Q^*(\bm{d}_n)$ is
\begin{equation}
    Q^*(\bm{d}_n) = \frac{1}{Z_D} \prod_{j=1}^J \prod_{l_i \in d_{nj}} \theta^*_{k_im_i}
\label{optqd}
\end{equation}
where $Z_D$ is a normalizing constant.

We can now obtain the optimal parameters $\theta^*_{k_im_i}$ using \eqref{thetastar}. Given that $Q^*(\bm{\theta})$ is a Dirichlet distribution, we have
\begin{align}
    \theta^*_{k_im_i} &=
    \text{exp}(\int_{\bm{\theta}} \textsc{Dirichlet}(\bm{\theta}_k; \bm{\omega}_k) \log \theta_{k_im_i}) \\
                  &= \text{exp}(\psi(\omega_{k_im_i}) - \psi(\sum_{m=1}^M \omega_{k_im}))  \label{digammas}
\end{align}
where $\psi$ is the digamma function.

\subsection{Variational Update Algorithm}
\label{sec:updates}

Given the expressions $Q^*(\bm{\theta})$ and $Q^*(\bm{d}_n)$, we can formulate a variational update algorithm to update $Q(\bm{\theta})$ and $Q(D)$ iteratively to a local maximum. First consider $Q^*(\bm{\theta})$, repeated in (\ref{optq}-\ref{optqw}).

\begin{equation}
        Q^*(\bm{\theta}) = \prod_{k=1}^{K} \prod_{m=1}^{M} \theta_{km}^{\omega_{km}-1}
        \label{optq}
\end{equation}
where
\begin{equation}
\omega_{km} = \alpha_{km} + \sum_{n=1}^N Q(\bm{d}_n) \sum_{j=1}^J \sum_{l_i \in d_{nj}} \delta(k_{i} = k, m_{i} = m) \label{optqw}
\end{equation}

Equations (\ref{optq}-\ref{optqw}) show that we can calculate $Q^*(\bm{\theta})$ by taking the prior pseudo-counts $\alpha_{km}$ and adding to them a term that depends on $Q(\bm{d}_n)$. Consider what the term we are adding to $\alpha_{km}$ represents. We sum over all possible derivations for the $N$ sentences in the corpus and take their probability. The delta term is 1 only for lexical items $l_{k_{i}m_{i}} = l_{km}$, so the sums over the delta terms count how many times $l_{km}$ shows up in the derivations in $D$. With this in mind, we can restate \eqref{optqw} as \eqref{optqw2}.
\begin{equation}
    \omega_{km} = \alpha_{km} +
    \sum_{n=1}^N \sum_{j=1}^J Q(d_{nj}|s_n) c(l_{km}; d_{nj})
    \label{optqw2}
\end{equation}
where $c(l; d)$ returns the amount of occurrences of a lexical item $l$ in derivation $d$. \eqref{optqw2} gives us the quantity that we use to update each parameter: the expected count of each lexical item with respect to $Q(D)$.

Similarly, equations (\ref{thetastar}--\ref{digammas}) show us how to update $Q(D)$. We simply compute the difference of digammas in \eqref{digammas} over $\bm{\omega}$, the updated parameters in $Q(\bm{\theta})$.

Using expressions \eqref{optqw2} and \eqref{digammas}, we state the variational update algorithm in (\ref{wzero}-\ref{thetai}). Let $q^{(i)}(\bm{\theta}, D)$ be the approximation of $Q(\bm{\theta}, D)$ at iteration $i$ of the algorithm. We alternate between (i) updating $q^{(i)}(\bm{\theta})$ via updating its parameters $\bm{\omega}^{(i)}$, and (ii) updating $q^{(i)}(D)$ via updating its parameters $\bm{\theta}^{(i)}$. In \eqref{wzero}, we initialize $\bm{\omega}^{(0)}$ to the prior parameters $\bm{\alpha}$.  Then we iterate over $i$ until $ELBO(P, q^{(i)})$ converges, updating parameters according to (\ref{wiplus1}--\ref{thetai}).

\begin{align}
    \omega_{km}^{(0)} &= \alpha_{km} \label{wzero} \\
    \omega_{km}^{(i+1)} &= \alpha_{km} +
    \sum_{n=1}^N \sum_{j=1}^J q^{(i)}(d_{nj}|s_n) c(l_{km}; d_{nj}) \label{wiplus1} \\
    q^{(i)}(d_{nj}|s_n) &= \frac{1}{Z_D} \prod_{j=1}^J \prod_{l_i \in d_{nj}} \theta^{(i)}_{k_im_i} \label{qi} \\
    \theta^{(i)}_{k_im_i} &= \text{exp}(\psi(\omega_{k_im_i}) - \psi(\sum_{m=1}^M \omega_{k_im}))
    \label{thetai}
\end{align}

\section{Conclusion}
In this paper, we presented a minimalist grammar formalization and an approach to the application of the variational Bayesian inference for minimalist grammar induction. In future work, we hope to implement the variational algorithm we derived to  build a working minimalist grammar induction system. We intend to integrate this technical work with the generalized parser developed in \citeA{harasietal17} as part of a technical framework for evaluating theories of grammar within the minimalist program. We hope that this work will be used to provide a baseline for future development of language learning models as well as more 'human'-like natural language processing applications.

\bibliographystyle{apacite}
\bibliography{MGinductiontechReport}
\end{document}